\pdfoutput=1
\documentclass[11pt]{article}

\usepackage[]{acl}

\usepackage{times}
\usepackage{latexsym}
\usepackage{booktabs}
\usepackage{multirow}
\usepackage{xcolor, soul}
\sethlcolor{green}
\usepackage[T1]{fontenc}
\usepackage[utf8]{inputenc}
\usepackage{graphicx}
\usepackage{subcaption}
\usepackage{paralist}
\usepackage{enumitem}
\usepackage{colortbl}
\usepackage{microtype}
\usepackage{amsmath}
\usepackage{xspace}
\usepackage[normalem]{ulem}
\usepackage{wrapfig}
\usepackage{tabularx}

\newcommand{\beep}{\textsc{BEEP}\xspace}

\newcolumntype{L}[1]{>{\raggedright\let\newline\\\arraybackslash\hspace{0pt}}p{#1}}
\newcolumntype{M}[1]{>{\raggedright\let\newline\\\arraybackslash\hspace{0pt}}m{#1}}

\definecolor{darkgreen}{rgb}{0,0.5,0.1}
\definecolor{betterblue}{rgb}{0,0.1,0.75}
\definecolor{betterred}{rgb}{0.7,0.2,0.2}
\newcommand{\colordg}[1]{\textcolor{darkgreen}{\bf#1}}
\newcommand{\colorblue}[1]{\textcolor{betterblue}{\bf#1}}
\newcommand{\colorred}[1]{\textcolor{betterred}{\bf#1}}

\title{Literature-Augmented Clinical Outcome Prediction}

\author{
    Aakanksha Naik \textsuperscript{1}\thanks{\hspace{0.2em} Work done during internship at AI2.}\quad
    Sravanthi Parasa\textsuperscript{2}\quad
    Sergey Feldman\textsuperscript{3}\quad
    Lucy Lu Wang\textsuperscript{3}\quad
    Tom Hope\textsuperscript{3,4}\\
  \textsuperscript{1}Language Technologies Institute, Carnegie Mellon University \\
  \textsuperscript{2}Swedish Medical Group \quad
  
  \textsuperscript{3}Allen Institute for Artificial Intelligence \\ \textsuperscript{4} Paul G. Allen School for Computer Science \& Engineering, University of Washington\\
  \texttt{anaik@andrew.cmu.edu}\quad \texttt{\{tomh,lucyw,sergey\}@allenai.org} \\ }

\begin{document}
\maketitle
\begin{abstract}
We present \beep (Biomedical Evidence-Enhanced Predictions), a novel approach for clinical outcome prediction that retrieves patient-specific medical literature and incorporates it into predictive models.\footnote{Our code is available at \href{https://github.com/allenai/BEEP}{https://github.com/allenai/BEEP}.} Based on each individual patient's clinical notes, we train language models (LMs) to find relevant papers and fuse them with information from notes to predict outcomes such as in-hospital mortality. We develop methods to retrieve literature based on noisy, information-dense patient notes, and to augment existing outcome prediction models with retrieved papers in a manner that maximizes predictive accuracy. Our approach boosts predictive performance on three important clinical tasks in comparison to strong recent LM baselines, increasing F1 by up to 5 points and precision@Top-K by a large margin of over 25\%.

\end{abstract}

\section{Introduction}
Predicting the medical outcomes of hospitalized patients holds the promise of enhancing clinical decision making. 
% With the advent of electronic health records (EHRs), training AI models for outcome prediction has gained traction \cite{rajkomar2018scalable,hashir2020towards}. 
With the advent of electronic health records (EHRs), more clinical data has become available to train AI models for outcome prediction \cite{rajkomar2018scalable,hashir2020towards}. In particular, language models pretrained on biomedical and/or clinical text are demonstrating increasing proficiency when fine-tuned for the task of predicting outcomes such as in-hospital mortality or length of stay \cite{van-aken-etal-2021-clinical}. 

% \lucy{Existing work has focused on using only clinical notes for outcome prediction \cite{boag2018s,hashir2020towards,van-aken-etal-2021-clinical}. Though \citet{van-aken-etal-2021-clinical} incorporate retrieved documents during pretraining as a source of additional knowledge, these documents are not used at test time. In this work, we explore a novel approach that dynamically retrieves relevant medical literature for each patient, and incorporates this literature to improve patient-specific prediction.}

In this work, we explore a novel approach for improving clinical outcome prediction by dynamically retrieving relevant medical literature for each patient, and incorporating this literature into language models (LMs) trained for outcome prediction from clinical notes. This is in contrast to existing outcome prediction work that uses \emph{only} clinical notes \cite{boag2018s,hashir2020towards}.  Recent LM-based approaches \citet{van-aken-etal-2021-clinical} have designed pretraining schemes over corpora of clinical notes and \emph{general} biomedical literature. This is in contrast to our work, where we directly incorporate a literature retrieval mechanism into our outcome prediction model, by finding papers relevant to \emph{specific} patient cases. 
%Recently, \citet{van-aken-etal-2021-clinical} incorporated biomedical literature during pretraining, but unlike our work, they do not tailor literature retrieval to specific patient cases or use it directly for outcome prediction. 
Our approach, named \beep (Biomedical Evidence-Enhanced Predictions), is broadly inspired by Evidence Based Medicine (EBM)---a leading paradigm in modern medical practice which calls for finding the ``current best evidence''
% to support clinical decision-making regarding individual patient care
to support optimal clinical decisions for each \emph{individual} patient \cite{sackett1996evidence}. 

% In this work, we explore a novel approach for improving clinical outcome prediction by dynamically retrieving relevant medical literature for each patient, in contrast to existing work on clinical prediction that only uses clinical notes \cite{boag2018s,hashir2020towards}. Recently, \citet{van-aken-etal-2021-clinical} incorporated biomedical literature during pretraining, but unlike our work, they do not tailor literature to patient cases or use it directly for outcome prediction. We explore approaches for retrieving literature pertaining to each patient's case, and develop methods for incorporating them into a clinical language model trained for outcome prediction. Our approach, named \beep (Biomedical Evidence-Enhanced Predictions), is broadly inspired by Evidence Based Medicine (EBM)---a leading paradigm in modern medical practice which calls for finding the ``current best evidence''
% % to support clinical decision-making regarding individual patient care
% to support optimal clinical decisions for each \emph{individual} patient \cite{sackett1996evidence}. 

\begin{figure}
    \centering
    \includegraphics[width=0.47\textwidth]{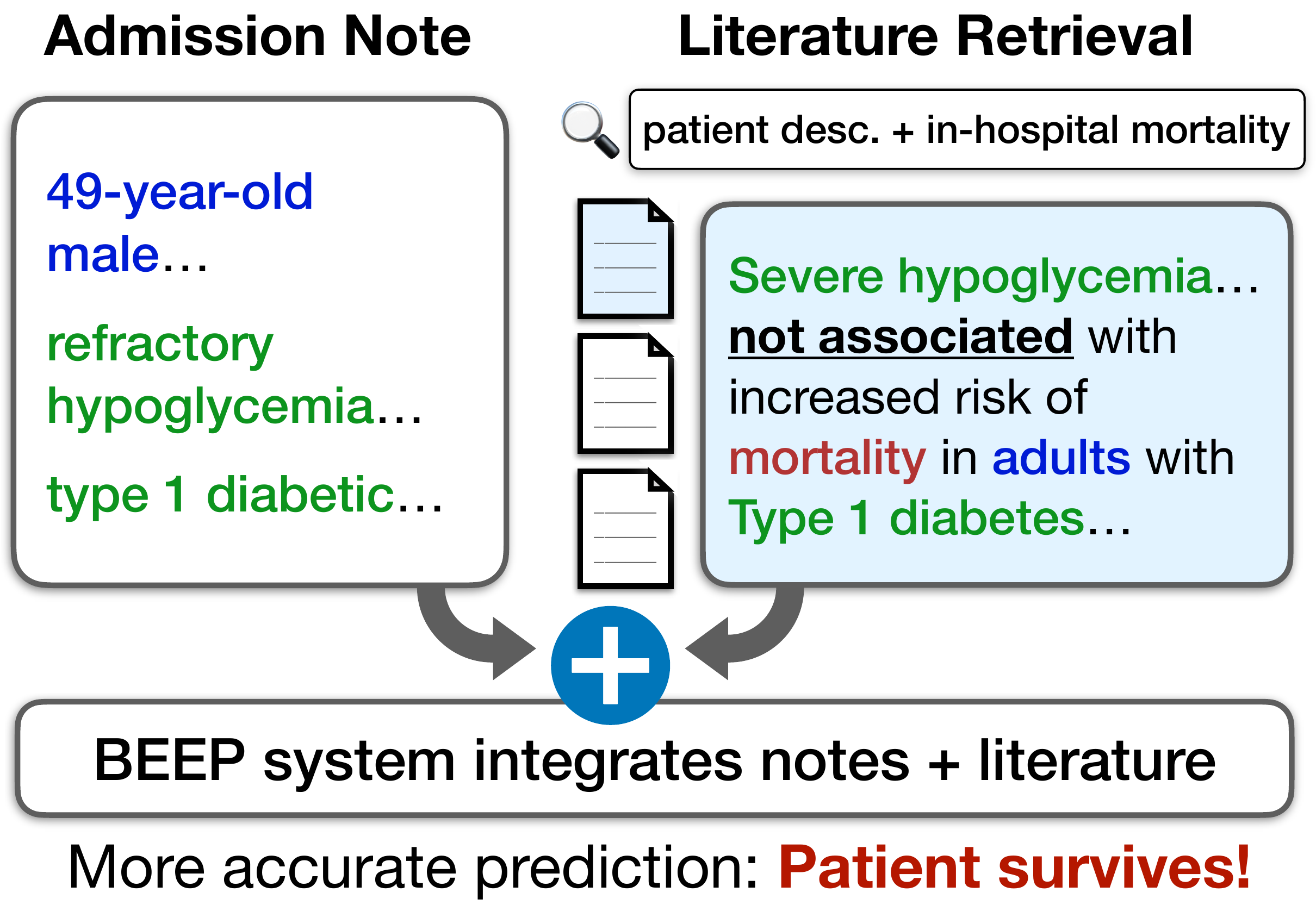}
    \vspace{-0.2em}
    \caption{Overview of \beep. We retrieve literature relevant to the patient description and an outcome of interest, in-hospital mortality in this example. We combine both sources of information to train a model to predict the outcome with better accuracy.}
    \label{fig:example}
    \vspace{-0.5em}
\end{figure}

Our setting presents unique challenges. First, our approach requires retrieving literature based on noisy EHR notes containing multitudes of information (e.g., medical history, ongoing treatments), unlike orthogonal efforts on extracting and summarizing scholarly information related to well-formed questions (e.g., the efficacy of ACE inhibitors in adult patients with type-2 diabetes) \cite{wallace2019does,lehman2019inferring,deyoung2020evidence,deyoung-etal-2021-ms}. In addition, as our end task is predicting patient outcomes, another challenge lies in aggregating the retrieved literature in a way that maximizes prediction accuracy. Toward these challenges, we make the following key contributions:

\begin{itemize}[leftmargin=1em,labelwidth=*,align=left,itemsep=1pt,parsep=1pt]

\item \textbf{Literature-Augmented Model.} As illustrated in Figure \ref{fig:example}, for each ICU patient and each target outcome to be predicted (e.g., mortality), our model retrieves papers from PubMed, encoded and fused together with the ICU admission note for making a final prediction. We present several architectures for retrieving papers and for aggregating and combining them with clinical notes. We make our code, cohort selection, paper identifiers and models publicly available.

\item \textbf{Adding Literature Boosts Results.} For evaluation, we measure both overall performance and precision/recall@Top-K, to account for the real-world scenario where ``alarms'' are only raised for high-confidence predictions to avoid alarm fatigue \cite{sendelbach2013alarm}. \beep provides substantial improvements over baselines, with strong gains in overall classification performance and precision@Top-K. For example, we improve F1 by up to 5 points and precision@Top-K by a large margin of over 25\%. 

%For example, AUROC for mortality improves by over 3 points; and top 10\% mortality predictions from our model yield a dramatic >25\% boost over the baseline's top 10\% precision. %Finally, we conduct empirical analysis to better understand the gains from using \beep.

\item \textbf{Exploring Patient-Specific Retrieval.} We explore a range of sparse and dense retrieval approaches, including language models, for the complex and underexplored task of retrieving relevant literature based on a patient's noisy, information-dense clinical note. Our final retrieval module employs a retrieve-rerank approach that effectively retrieves helpful literature, as shown in our analysis (section~\ref{sec:analysis}).

\end{itemize}

We hope our work opens new research directions for automatically scanning literature for patient-specific evidence, and combining it with EHR information to boost accuracy of medical predictive models. Finally, our work raises the more general prospect of building predictive models that can dynamically \emph{learn to retrieve literature} for optimizing task accuracy, in medicine and other related areas. 

%Additionally, we benchmark several recent clinical language model baselines, including an LM that uses a novel pre-training objective designed for clinical predictions \cite{van-aken-etal-2021-clinical}, and an LM that integrates external biomedical knowledge and has not been evaluated outside intrinsic tasks like NER \cite{michalopoulos-etal-2021-umlsbert}. 

%Interestingly, we find the latter LM, which incorporates biomedical knowledge, to outperform the model with specialized task-specific pre-training.
\section{Related Work}
\textbf{Patient-Specific Literature Retrieval.} %Given the existing volume of biomedical literature and the rapid pace at which new research is published \cite{hunter2006biomedical}, automatic retrieval of literature relevant to specific patients can be extremely informative and time-saving for clinical practitioners. 
Since 2014, the Text REtrieval Conference (TREC) has organized a series of challenges to advance research in this area. The TREC Clinical Decision Support (CDS) tracks focused on evaluating systems on the task of retrieving biomedical articles relevant for answering generic clinical questions about patient medical records (e.g., identifying potential diagnoses, treatments, and tests) \cite{simpson2014overview,roberts2015overview,DBLP:conf/trec/RobertsDVH16}. TREC CDS 2014 and 2015 used short case reports as idealized representations of medical records due to the lack of available de-identified records. TREC 2016 shifted to using real-world medical records from the Medical Information Mart for Intensive Care (MIMIC) database \cite{johnson2016mimic}.\footnote{Since 2017, the focus has switched to TREC-PM (precision medicine) tracks where articles are retrieved based on short structured queries with attributes such as patient condition and demographics, a less realistic scenario.} In our work, our focus is on \emph{predicting clinical outcomes} using ICU admission notes and patient-specific retrieved literature. 

%due to the mismatch in language between case reports and noisier records. 
%The TREC 2016 formulation comes closest to our system, which also retrieves articles relevant to medical records, though 
% Since 2017, TREC has switched its focus to precision medicine, with the TREC-PM tracks aiming to evaluate systems on the task of retrieving biomedical articles according to short, structured queries containing attributes such as patient condition and demographics \cite{DBLP:conf/trec/RobertsDVHBLP17,DBLP:conf/trec/RobertsDVHBL18,DBLP:conf/trec/RobertsDVHBLPM19,DBLP:conf/trec/RobertsDVBH20}, a less realistic setting than TREC CDS 2016.

%The TREC-CDS challenges spurred the development of several retrieval models for this task, including neural approaches \cite{balaneshin2016optimization,goodwin2017knowledge,ran2017document,alsulmi2018improving,das-etal-2020-sequence}. 
%Few studies have explored the utility of large pretrained language models for these tasks.
\newcite{alberto2021structured} use contextualized representations on more structured retrieval tasks not involving clinical notes \cite{voorhees2021trec}, leaving open the question of how large pretrained language models (LMs) would fare on long, noisy EHR text. 
% Unlike existing work,
We explore this by experimenting with LMs for retrieval based on EHR text.\\ [-4mm]
% retrieval models.

\begin{figure*}
    \centering
    \includegraphics[width=\linewidth]{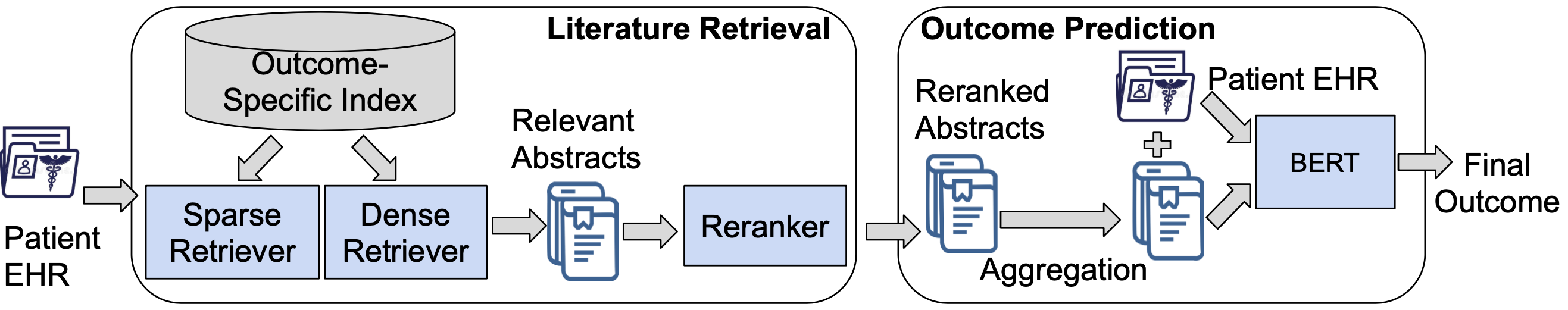}
    \caption{Complete system pipeline, unpacking the high-level overview seen in Figure \ref{fig:example}. For a given patient ICU admission note, the literature retrieval module first retrieves relevant biomedical abstracts from a clinical outcome-specific index, then reranks a top-ranked subset of abstracts. The outcome prediction module aggregates information from these reranked abstracts and fuses it with the admission note to make the final prediction}
    \label{fig:pipeline}
    \vspace{-0.5em}
\end{figure*}

% \noindent
% \textbf{Evidence Synthesis:}\\
\noindent
\textbf{Clinical Outcome Prediction.} 
%The transition to Electronic Health Records (EHRs) and the creation of the MIMIC database \cite{saeed2002mimic,saeed2011multiparameter,johnson2016mimic} has encouraged the development of models that can analyze structured (e.g., lab results, vital signs, etc.) and unstructured (e.g., nursing notes) information from records to predict various clinical outcomes of interest. 
The idea of using automated outcome prediction for assisting clinical triage, workflow optimization, and hospital resource management has received much interest recently, especially given the conditions of the COVID-19 pandemic \cite{li2020automated}.
Predictive models based on structured (e.g., lab results) and unstructured (e.g., nursing notes) information have been built for 
key clinical outcomes including mortality \cite{jain2019analysis,feng-etal-2020-explainable}, length of hospital stay \cite{van-aken-etal-2021-clinical}, readmission \cite{jain2019analysis}, sepsis \cite{feng-etal-2020-explainable}, prolonged mechanical ventilation \cite{huang-etal-2020-clinical}, and diagnostic coding \cite{jain2019analysis,van-aken-etal-2021-clinical}. Increasingly, models have leveraged unstructured text from notes since they can contain key information for outcome prediction \cite{boag2018s,jin2018improving}. %With the advent of several large pretrained language models for clinical and biomedical tasks \cite{alsentzer-etal-2019-publicly,peng2019transfer,lee2020biobert,gu2021domain,van-aken-etal-2021-clinical,michalopoulos-etal-2021-umlsbert}, 
% This has been done recently using LMs 
Most recently, \citet{van-aken-etal-2021-clinical} attempted this using large pretrained LMs. 
Our work compares the performance of a broader range of state-of-the-art pretrained language models on outcome prediction tasks.%, and incorporate the best-performing models as base text encoders in \beep.

\section{\beep: Literature-Enhanced Clinical Predictive System}
\noindent
\textbf{Task \& Approach Overview.} Our goal is to improve models for clinical outcome prediction from EHR notes by augmenting them with relevant biomedical literature. \beep consists of two main stages: (i) literature retrieval, and (ii) outcome prediction. We also briefly experiment with a formulation that trains both jointly (details in section~\ref{sec:exp}). Given a patient EHR note $Q$ and a clinical outcome of interest $y$, the first stage is to identify a set of biomedical abstracts $Docs(Q) = \{D_1,...,D_n\}$ from PubMed\footnote{\href{https://pubmed.ncbi.nlm.nih.gov}{https://pubmed.ncbi.nlm.nih.gov}} that may be helpful in assessing the likelihood of the patient having that outcome. The next stage is to augment the input to an EHR-based outcome prediction model with these retrieved abstracts ($Q \cup Docs(Q))$ and predict the final outcome. Figure~\ref{fig:example} provides a high-level illustration of \beep, and Figure~\ref{fig:pipeline} unpacks it with more detail. Next, we describe our system's main components.
% The literature retrieval module takes as input a patient's clinical note and returns abstracts relevant to the patient and outcome of interest  The outcome prediction module aggregates these abstracts (i.e. evidence) and tries to predict the final clinical outcome using both the patient note and available evidence. 

\subsection{Literature Retrieval Module}
Our literature retrieval module consists of three components: (i) an index of biomedical abstracts pertaining to the outcome of interest, (ii) a \textit{retriever} that retrieves a ranked list of abstracts relevant to the patient note from the index, and (iii) a \textit{reranker} that reranks 
% top-ranked abstracts from the retriever
retrieved abstracts using a stronger document similarity computation model. For the retriever, we experiment with both sparse and dense models. We follow the standard retrieve-rerank approach, which has been shown to achieve good balance between efficiency and retrieval performance \cite{dang2013two}, and has recently also proved useful for large-scale biomedical literature search \cite{wang2021domain}. In the retrieval step, we prioritize efficiency, using models that scale well to large document collections but are not as accurate, to return a set of top documents. In the reranker step, we prioritize retrieval performance by running a computationally expensive but more accurate model on the 
% selected 
smaller set of retrieved documents.

\subsubsection{Outcome-Specific Index Construction}
Since we are interested in identifying information related to a \emph{specific outcome} for a patient, we begin by constructing an index of all abstracts from PubMed relevant to that outcome to limit search scope. To gather all abstracts relevant to a clinical outcome, we first identify MeSH (Medical Subject Heading) terms associated with the outcome by performing MeSH linking on the outcome descriptions using scispaCy \cite{neumann-etal-2019-scispacy}. These associated MeSH terms are then used as queries to retrieve abstracts.\footnote{\href{https://www.ncbi.nlm.nih.gov/books/NBK25499/}{https://www.ncbi.nlm.nih.gov/books/NBK25499/}} For some MeSH terms that are too broad (e.g., ``mortality''), we include additional qualifiers (e.g., ``human'') to make sure we do not gather articles that are not relevant to our overall patient cohort. Appendix~\ref{sec:pubquery} lists the final set of queries used for all clinical outcomes considered in this work. Abstracts retrieved via this process are used to construct the outcome-specific index.

\subsubsection{Sparse Retrieval Model}
The sparse retrieval model returns top-ranked abstracts based on cosine similarity between TF-IDF vectors of MeSH terms for the query (clinical note) and the documents (outcome-specific abstracts). MeSH terms from abstracts are extracted by running scispaCy MeSH linking over the abstract text. PubMed MeSH tagging is done only at the abstract level, and does not reflect actual term frequency in the text, requiring our extraction step. However, extracting MeSH terms from clinical notes requires a more elaborate pipeline, due to two major issues:
\begin{itemize}[leftmargin=1em,labelwidth=*,align=left,noitemsep,parsep=1pt]
    \item \textbf{Entity type and boundary issues:} Off-the-shelf entity extractors like scispaCy and cTAKES \cite{savova2010mayo} extract some entity types that are uninformative for relevant literature retrieval, e.g., hospital names, references to family members, etc. They also have a tendency to ignore important qualifiers. For example, given a sentence containing the entity ``right lower extremity pain'', both extractors returned ``extremity'' and ``pain'' as separate entities. 
    \item \textbf{Negated entities:} Clinical notes have a high density of negated entities (up to 50\% of \cite{chapman2001evaluation}). These entities must be identified and discarded prior to literature retrieval to avoid retrieving articles about symptoms and conditions that are \textit{not} exhibited by the patient.
\end{itemize}

% To handle these issues, we extract MeSH terms from clinical notes using the following process:\\ [-4mm]
To handle these issues, we train an entity extraction model that focuses on problems, tests, and treatments with empirically good coverage of important qualifiers \cite{uzuner20112010}. We then filter negated entities with negation detection \cite{harkema2009context} and perform entity linking to MeSH terms. For more information and implementation details see Appendix~\ref{sec:imp}.

% \noindent \textbf{Entity Extraction.} First, we extract entities from clinical notes using a model trained on the i2b2 2010 concept extraction dataset \cite{uzuner20112010}. This dataset consists of clinical notes annotated with three types of entities: problems, tests, and treatments. These entity types cover the pertinent medical information that can be used to retrieve abstracts relevant to a clinical note. Moreover, the i2b2 guidelines require annotators to include all qualifiers within an entity span, so training a model on these annotations should bias it towards including pertinent entity qualifiers. Our entity extraction model uses a BERT-based language model to compute token representations, followed by a linear layer to predict entity labels.\\ [-4mm]

% \noindent \textbf{Entity Filtering.} After extracting entities, we filter out all negated entities. Negated entities are detected using the ConText algorithm for negation detection from clinical text \cite{harkema2009context}.\\ [-4mm]

% \noindent \textbf{MeSH Linking.} Finally, the set of filtered entities is linked to MeSH terms using scispaCy. Entities not linked to MeSH terms are discarded. MeSH terms linked in clinical notes and abstracts are used to compute TF-IDF vectors for the sparse retrieval model.
% Implementation details for all models are provided in Appendix~\ref{sec:imp}.

\subsubsection{Dense Retrieval Model}
%One well-known issue with sparse retrievers is their inability to go beyond surface forms while computing query-document similarity. 
We add a dense retrieval model to complement the sparse retriever, an approach that has shown promise in recent work \cite{gao2021complement}. Our dense retrieval model maps clinical notes (queries) and biomedical abstracts (documents) to a shared dense low-dimensional embedding space. Computing similarity between these encoded vectors allows for softer matching beyond surface form. For dense retrieval, we use a BERT-based bi-encoder model. We use a bi-encoder to support scaling to large document collections, as opposed to cross-encoder models which are much slower (e.g., \cite{gu2021domain}). We use PubmedBERT \cite{gu2021domain} as the encoder and train our bi-encoder using the dataset from the TREC 2016 clinical decision support task \cite{DBLP:conf/trec/RobertsDVH16}. For more details, see Appendix~\ref{sec:imp}. Our bi-encoder achieves mean precision@10 score of 45.67 on TREC 2016 data in 5-fold cross-validation, comparable to state-of-the-art results \cite{das-etal-2020-sequence}.

\subsubsection{Reranker Model}
\label{sec:reranker_model}
The reranker model takes a subset of top-ranked documents from both the sparse and dense retrieval models and rescores them. We use a BERT-based cross-encoder model for reranking, prioritizing ranking performance over efficiency on this smaller subset. Given a query clinical note $Q$ and an abstract document $D_i$, we run a PubmedBERT-based encoder over the concatenation of both (\texttt{[CLS]} $Q$ \texttt{[SEP]} $D_i$ \texttt{[SEP]}) to compute an embedding $E_{QD_i}$. This embedding is run through a linear layer to produce a relevance score, trained using cross-entropy loss with respect to document relevance labels from the TREC 2016 dataset. Our cross-encoder achieves a mean precision@10 score of 48.33 on TREC 2016 in 5-fold cross-validation, which is also comparable to state-of-the-art performance on TREC CDS 2016 \cite{das-etal-2020-sequence}.

From the top-ranked documents returned by the reranker, the top $k$ are selected\footnote{We treat $k$ as a hyperparameter, see appendix~\ref{sec:tuning}. 
} to be passed alongside the patient clinical note to the outcome prediction module, which we describe next.

\subsection{Outcome Prediction Module}
The goal of this module is to compute an aggregate representation from the set of top $k$ abstracts relevant to the clinical note, and then predict the outcome of interest using this aggregate representation and the note representation. 

\subsubsection{Aggregation Strategies}
Let $Docs(Q) = {D_1,...,D_k}$ be the set of relevant abstracts retrieved for clinical note $Q$ and $\textsc{BERT}(X)$ be the encoder function that returns an embedding $E_X$ given a document $X$. We experiment with four different strategies to compute an aggregate literature representation for $Docs(Q)$, which we denote by $LR(Q)$.\\ [-4mm]

\noindent 
\textbf{Averaging.} Averaging encoder representations:
\begin{equation}
\setlength{\abovedisplayskip}{0pt}
    LR(Q) = \frac{1}{k} \sum_{i=1}^k \textsc{BERT}(D_i)
\end{equation}
\noindent
\textbf{Weighted Averaging.} Weighted average of encoder representations:
\begin{equation}
\setlength{\abovedisplayskip}{0pt}
    LR(Q) = \frac{1}{\sum_{i=1}^k w_i} \sum_{i=1}^k w_i \cdot \textsc{BERT}(D_i)
\end{equation}
where weights $w_i$ are the relevance scores computed by the reranker. The final outcome is computed by concatenating note representation $\textsc{BERT}(Q)$ with $LR(Q)$ and running this through a linear layer.\\
\begin{table}
\begin{subtable}{1\columnwidth}
    \centering
    \begin{tabular}{ccccc}
       \toprule \textbf{Outcome}  & \textbf{0} & \textbf{1} & \textbf{2} & \textbf{3}  \\ \midrule
        \textbf{PMV} & 3,776 & 3,335 & - & -\\
        \textbf{MOR} & 43,609 & 5,136 & - & - \\
        \textbf{LOS} & 5,596 & 16,134 & 13,391 & 8,488 \\ \bottomrule
    \end{tabular}
    \caption{Class distribution for all outcomes. For PMV, classes 0 and 1 refer to cases that don't/do require prolonged ventilation. For MOR, classes 0 and 1 refer to patients that don't/do die in admission. For LOS, classes 0-3 refer to stay lengths of <3 days, 3-7 days, 1-2 weeks, and >2 weeks respectively.}
    \vspace{1mm}
    \label{tab:data}
\end{subtable}
\begin{subtable}{1\columnwidth}
    \centering
    \begin{tabular}{ccccc}
       \toprule \textbf{Outcome}  & \textbf{Train} & \textbf{Dev} & \textbf{Test} & \textbf{\#Articles} \\ \midrule
        \textbf{PMV} & 5,691 & 712 & 708 & 81,311\\
        \textbf{MOR} & 33,997 & 4,918 & 9,830 & 90,125\\
        \textbf{LOS} & 30,421 & 4,391 & 8,797 & 93,594\\ \bottomrule
    \end{tabular}
    \caption{Training, development and test splits, and total number of PubMed articles in our outcome-specific index for each clinical outcome.}
    \label{tab:datasplits}
    \vspace{-0.5em}
\end{subtable}

\caption{Data statistics per outcome}
\vspace{-0.6em}
% \begin{subtable}{1\columnwidth}
%     \centering
%     \begin{tabular}{cc}
%     \toprule \textbf{Outcome}& \textbf{\#Articles} \\ \midrule
%     \textbf{PMV} & 81,311\\
%     \textbf{MOR} & 90,125\\
%     \textbf{LOS} & 93,594\\ \bottomrule
%     \end{tabular}
%     \caption{Total number of PubMed articles for each clinical outcome contained in our outcome-specific index.}
%     \label{tab:litcount}
% \end{subtable}
\end{table}
We also concatenate the note embedding with each abstract ($E_{QD_i} = [\textsc{BERT}(Q);\textsc{BERT}(D_i)]$), run outcome prediction and aggregate output probabilities as follows.

\noindent
\textbf{Soft Voting.} Averaging per-class probabilities from $k$ outcome prediction runs:
\begin{equation}
\setlength{\abovedisplayskip}{0pt}
\setlength{\belowdisplayskip}{0pt}
    p(y=c) = \frac{1}{k} \sum_{i=1}^k p(y=c | E_{QD_i})
\end{equation}

\noindent
\textbf{Weighted Voting.} Weighted average of per-class probabilities from $k$ outcome predictions runs:
\begin{equation}
\setlength{\abovedisplayskip}{0pt}
\setlength{\belowdisplayskip}{0pt}
    p(y=c) = \frac{1}{\sum_{i=1}^k w_i} \sum_{i=1}^k w_i \cdot p(y=c | E_{QD_i})
\end{equation}
\begin{table*}[]
    \centering
    \small
    \setlength{\tabcolsep}{4.1pt}
    \begin{tabular}{lccccccccc}
    \toprule  & \multicolumn{3}{c}{PMV} & \multicolumn{3}{c}{MOR} & \multicolumn{3}{c}{LOS} \\ \cmidrule{2-10}
    \textbf{Model} & \textbf{AUROC} & \textbf{Micro F1} & \textbf{Macro F1} & \textbf{AUROC} & \textbf{Micro F1} & \textbf{Macro F1} & \textbf{AUROC} & \textbf{Micro F1} & \textbf{Macro F1} \\ \midrule
    \textbf{BLUEBERT} & 54.27 & 53.25 & 51.64 & 81.49 & 89.11 & \textbf{62.69} & \underline{\textbf{73.22}} & 45.66 & \underline{\textbf{44.18}}\\
    \textbf{\qquad+Avg} & 57.21 & 54.66 & 52.32 & 83.90 & 90.52 & 61.62 & 71.66 & 45.22 & 40.66\\
    \textbf{\qquad+SVote} & \textbf{58.16} & \textbf{56.07} & 52.63 & 84.21 & \textbf{90.60} &  61.00  & 72.54 & \textbf{46.02} & 42.46\\
    \textbf{\qquad+WVote} & 57.71 & 57.91 & \underline{\textbf{56.67}} & 84.00 & 90.45 & 61.02 & 71.49 & 44.82 & 39.55\\
    \textbf{\qquad+WAvg} & 57.59 & 55.65 & 52.21 & \textbf{84.26} & 90.44 & 60.49 & 72.58 & 45.90 & 42.39\\ \midrule
    \textbf{UMLSBERT} & 56.44 & 56.07 & \textbf{54.97} & 83.34 & 87.93 & 66.93 & 72.19 & 43.12 & 42.20 \\
    \textbf{\qquad+Avg} & 58.36 & \underline{\textbf{56.50 }}& 54.62 & 84.02 & 90.41 & 60.28 & 72.25 & 45.61 & 41.58\\
    \textbf{\qquad+SVote} & 55.92 & 54.66 & 50.94 & 83.30 & 84.82 & \underline{\textbf{67.23}} & 72.14 & 45.55 & 42.12\\
    \textbf{\qquad+WVote} & \underline{\textbf{59.43}} & 56.07 & 54.26 & \underline{\textbf{84.65}} & \underline{\textbf{90.62}} & 62.93 & \textbf{72.71} & \underline{\textbf{46.44}} & \textbf{42.71}\\
    \textbf{\qquad+WAvg} & 59.30 & \underline{\textbf{56.50}} & 53.70 & 83.59 & 90.35 & 59.61 & 71.02 & 44.58 & 39.95\\
\bottomrule
    \end{tabular}
    \caption{Performance of baseline and literature-augmented outcome prediction models on all clinical outcomes. We note that LOS is a multiclass target; we observe substantial gains in 2/4 of the classes (Table~\ref{tab:lostop} in the Appendix).}
    \label{tab:results}
\end{table*}
\vspace{-2em}
\section{Experiments \& Results}
\label{sec:exp}
We test our system on the task of predicting clinical outcomes from patient admission notes. Predicting outcomes from admission notes can 
% be extremely insightful in 
help with early identification of 
% potential risk
at-risk patients and assist hospitals in resource planning by indicating how long patients may require hospital/ICU beds, ventilators etc. \cite{van-aken-etal-2021-clinical}.

% \begin{table}[]
%     \centering
%     \begin{tabular}{ccccc}
%       \toprule \textbf{Outcome}  & \textbf{No PMV} & \textbf{PMV} & & \\ \midrule
%         \textbf{PMV} & 3,776 & 3,335 & \cellcolor[gray]{0.8} & \cellcolor[gray]{0.8}\\
%         \toprule \textbf{Outcome}  & \textbf{No MOR} & \textbf{MOR} & & \\ \midrule
%         \textbf{MOR} & 43,609 & 5,136 & \cellcolor[gray]{0.8} & \cellcolor[gray]{0.8} \\
%         \toprule \textbf{Outcome}  & \textbf{<3 days} & \textbf{>=3 and <=7 days} & \textbf{>7 and <=14 days} & \textbf{>14 days}\\ \midrule
%         \textbf{LOS} & 5,596 & 16,134 & 13,391 & 8,488 \\ \bottomrule
%     \end{tabular}
%     \caption{Class distribution for all clinical outcomes}
%     \label{tab:data}
% \end{table}

\subsection{Clinical Outcomes}
We evaluate our system on three clinical outcomes: 
\begin{itemize}[leftmargin=1em,labelwidth=*,align=left,itemsep=1pt,parsep=1pt]
    \item \textbf{PMV:} Prolonged mechanical ventilation prediction, identifying whether a patient will require ventilation for >7 days \cite{huang-etal-2020-clinical}.
    \item \textbf{MOR:} In-hospital mortality prediction, identifying whether a patient will survive their current admission \cite{van-aken-etal-2021-clinical}.
    \item \textbf{LOS:} Length of stay prediction is the task of identifying how long a patient will need to stay in the hospital. We follow \citet{van-aken-etal-2021-clinical} and group patients into four major categories based on clinician recommendations: <3 days, 3-7 days, 1-2 weeks, and >2 weeks. 
\end{itemize}

PMV and MOR are binary classification tasks, while LOS is a multi-class classification task. We predict these outcomes from patient admission notes extracted from the MIMIC III v1.4 database \cite{johnson2016mimic}, which contains de-identified EHR data including clinical notes in English from the Intensive Care Unit (ICU) of the Beth Israel Deaconess Medical Center in Massachusetts between 2001 and 2012. Admission notes are constructed by filtering discharge summary documents from MIMIC to only retain the following sections typically known at admission: Chief complaint, (History of) Present illness, Medical history, Admission medications, Allergies, Physical exam, Family history and Social history. Notes that do not contain any of these sections are excluded. For PMV, we follow the cohort selection process from \newcite{huang-etal-2020-clinical}, and include all patients who were above 18 years of age and were on mechanical ventilation for at least 2 days with more than 6 hours each day. Patients transferred from other hospitals, organ donors, and patients with neuromuscular disease, head and neck cancer, and extensive burns, which always lead to PMV and may act as confounds, were excluded. For MOR and LOS, we follow the same cohort selection process as \newcite{van-aken-etal-2021-clinical}, and include all patients except newborns and remove duplicate admissions. Following these cohort selection processes results in the data splits shown in Table~\ref{tab:datasplits}. Table~\ref{tab:datasplits} also shows the numbers of relevant PubMed articles for all three clinical outcomes.

\subsection{Selecting the Encoder Language Model}
Since the encoder used for outcome prediction needs to produce representations for both clinical notes and relevant abstracts, we 
% need to choose BERT-based language models 
choose language models
that have been pretrained on \emph{both} biomedical and clinical text. We evaluate the following models on outcome prediction (without literature augmentation) to choose a suitable encoder:

\begin{itemize}[leftmargin=1em,labelwidth=*,align=left,itemsep=1pt,parsep=1pt]
    \item \textbf{ClinicalBERT} \cite{alsentzer-etal-2019-publicly}: ClinicalBERT further pretrains BioBERT \cite{lee2020biobert}, a biomedical language model, on EHR notes from MIMIC III. We evaluate both versions: one trained on discharge summary notes only, and one trained on both discharge summaries and nursing notes.
    \item \textbf{CORe} \cite{van-aken-etal-2021-clinical}: CORe further pretrains BioBERT with a next sentence prediction objective on sentences describing admissions and outcomes. CORe jointly trains on EHR notes and biomedical articles.
    \item \textbf{BLUEBERT} \cite{peng2019transfer}: BLUEBERT further pretrains BERT \cite{devlin-etal-2019-bert} jointly on EHR notes and PubMed abstracts.
    \item \textbf{UMLSBERT} \cite{michalopoulos-etal-2021-umlsbert}: UMLSBERT further pretrains ClinicalBERT on EHR notes from MIMIC, with tweaks to the architecture and pretraining objective to incorporate conceptual knowledge from the Unified Medical Language System (UMLS) Metathesaurus \cite{schuyler1993umls}.
\end{itemize}

Note that in this experiment, we predict clinical outcomes from patient admission notes only, without incorporating literature. We also use weighted cross-entropy loss to manage class imbalance (see Appendix~\ref{sec:imp}). Table~\ref{tab:lm} in the Appendix shows the performance of the above language models on the validation sets for all clinical outcomes. We select the top-performing language models BLUEBERT and UMLSBERT for our remaining experiments.\footnote{We also experiment with CORe but observe consistently lower scores (Table~\ref{tab:resultscore} in Appendix~\ref{sec:core}).}

% (i) prolonged mechanical ventilation, (ii) in-hospital mortality, and (iii) length of hospital stay. We use the  
\subsection{Literature Augmentation Results}

We provide two sets of results: for overall performance, and for high-confidence predictions. 

\paragraph{Overall Performance.} Table~\ref{tab:results} shows the overall performance of our literature-augmented outcome prediction system on all three clinical outcomes. We test our system using both UMLSBERT and BLUEBERT as encoders, as well as all four literature aggregation strategies. We report three metrics for each setting: (i) area under the receiver operating characteristic (AUROC), (ii) micro-averaged F1 score, and (iii) macro-averaged F1 score. From Table~\ref{tab:results}, we observe that incorporating literature leads to performance improvements on two of three clinical outcomes, PMV and mortality. On LOS prediction, results are more mixed, with minor improvements on micro F1 but no improvements on other metrics. Comparing BLUEBERT and UMLSBERT, variants that use UMLSBERT do slightly better on PMV and mortality, while results on LOS are more mixed. Comparing across literature aggregation strategies, there is no clear winner, though voting-based strategies seem to have a slight advantage, especially on UMLSBERT.

\begin{table}[t]
\begin{subtable}{1\columnwidth}
\small
\scalebox{0.9}{
    \centering
    \begin{tabular}{lcccc}
     \toprule & \multicolumn{2}{c}{\textbf{No PMV}} & \multicolumn{2}{c}{\textbf{PMV}}\\ \cmidrule{2-5}
     \textbf{Model} & \textbf{Prec@10} & \textbf{Rec@10} & \textbf{Prec@10} & \textbf{Rec@10}\\ \midrule
    \textbf{BLUEBERT} & 52.86 & 9.95 & 55.71 & 11.61 \\ %& 27.86\\
    \textbf{\qquad+Avg} & \textbf{64.29} & \textbf{12.1} & 60.0 & 12.5 \\ %& 30.00\\
    \textbf{\qquad+SVote} & 61.43 & 11.56 & \textbf{64.29} & \textbf{13.39} \\ %& 32.14\\
    \textbf{\qquad+WVote} & 62.86 & 11.83 & 52.86 & 11.01 \\ %& 26.43\\
    \textbf{\qquad+WAvg} & 58.57 & 11.02 & 52.86 & 11.01 \\ \midrule %& 26.43\\ 
    \textbf{UMLSBERT} & 58.57 & 11.02 & 57.14 & 11.90 \\ %& 28.57 \\
    \textbf{\qquad+Avg} & 67.14 & 12.63 & \textbf{64.29} & \textbf{13.39} \\ %& 32.14\\
    \textbf{\qquad+SVote} & 61.43 & 11.56 & 62.86 & 13.1 \\ %& 31.43\\
    \textbf{\qquad+WVote} & 64.29 & 12.1 & \textbf{64.29} & \textbf{13.39} \\ %& 32.14\\
    \textbf{\qquad+WAvg} & \textbf{68.57} & \textbf{12.9} & 62.86 & 13.1 \\ %& 31.43\\
\bottomrule     
    \end{tabular}
}
    \caption{For PMV}
    \label{tab:pmvtop}
\end{subtable}

\begin{subtable}{1\columnwidth} 
\small
\scalebox{0.9}{
    \centering
    \begin{tabular}{lcccc}
     \toprule & \multicolumn{2}{c}{\textbf{No MOR}} & \multicolumn{2}{c}{\textbf{MOR}} \\ \cmidrule{2-5}
     \textbf{Model} & \textbf{Prec@10} & \textbf{Rec@10} & \textbf{Prec@10} & \textbf{Rec@10} \\ \midrule
    \textbf{BLUEBERT} & \textbf{99.8} & \textbf{11.15} & 46.39 & \textbf{23.62} \\
    \textbf{\qquad+Avg} & 99.59 & 11.13 & 68.91 & 17.81 \\
    \textbf{\qquad+SVote} & 99.69 & 11.14 & \textbf{73.39} & 16.55  \\
    \textbf{\qquad+WVote} & 99.59 & 11.13 & 68.36 & 16.94 \\
    \textbf{\qquad+WAvg} & \textbf{99.8} & \textbf{11.15} & 69.46 & 16.07 \\ \midrule
    \textbf{UMLSBERT} & 99.8 & 11.15 & 42.06 & \textbf{39.21}\\
    \textbf{\qquad+Avg} & 99.59 & 11.13 & \textbf{69.07} & 15.78 \\
    \textbf{\qquad+SVote} & 99.8 & 11.15 & 40.69 & 38.72 \\
    \textbf{\qquad+WVote} & 99.49 & 11.12 & 68.44 & 19.94 \\
    \textbf{\qquad+WAvg} & \textbf{100.0} & \textbf{11.17} & 68.92 & 14.81 \\
    \bottomrule     
    \end{tabular}
}
\caption{For MOR}
\label{tab:mortop}
\vspace{-0.6em}
\end{subtable}
\caption{Precision and recall scores for top 10\% high-confidence predictions per class.}
\vspace{-0.7em}
\end{table}
\paragraph{Evaluating High-Confidence Predictions.}
In addition to standard evaluation, we evaluate the top 10\% high-confidence predictions per class for all models (precision/recall@TOP-K), informative for two key reasons. First, when using automated outcome prediction systems in a clinical setting, it is reasonable to only consider raising alarms for high-confidence positive predictions to avoid alarm fatigue \cite{sendelbach2013alarm}. Second, high-confidence predictions for both positive and negative classes can be used to reliably assist with hospital resource management (e.g., predicting future ventilation and hospital bed needs). %The idea of improving automated outcome prediction with the goal of assisting clinical triage, workflow optimization and hospital resource management has received much interest recently, given the conditions during the COVID-19 pandemic \cite{li2020automated}. Therefore, we evaluate the performance difference that adding literature provides, when considering per-class high-confidence predictions only.

Tables~\ref{tab:pmvtop} and \ref{tab:lostop} show the precision/recall\-@TOP-K scores for all models on prolonged mechanical ventilation, mortality, and length of stay prediction. In Table~\ref{tab:pmvtop}, we see that our literature-augmented models achieve much higher precision scores than the baseline ($\sim$9-12 points higher in most cases) for the PMV negative class. We also see higher precision scores than the baseline for the positive class ($\sim$5-9 points higher in most cases). This is a strong indicator that our literature-augmented pipeline might offer more utility for PMV detection in a clinical setting than using EHR notes only. Table~\ref{tab:mortop} shows similarly encouraging trends for mortality prediction. The mortality prediction dataset is the most skewed of the three datasets, and therefore we do not see much performance difference across models on the negative class. However, on the positive class, our literature-augmented models show dramatic increase in precision. In particular, BLUEBERT-based literature models show an increase in precision of $\sim$22-27 points, at the expense of only $\sim$6-7 point drop in recall relatively to non-literature models.\footnote{Note that since the MOR class is rare, a larger recall drop could still translate to a small number of incorrect cases only} This also indicates that literature-augmented mortality prediction might be more precise and reliable in a clinical setting than using clinical notes alone. From Table~\ref{tab:lostop} (Appendix \ref{sec:highconfinc}), we can see that for LOS prediction, our models show clear gains ($\sim$2-5 points) on classes 1 and 2 (i.e., 3-7 days and 1-2 weeks), and minor gains for some variants on class 3 (>2 weeks). We also perform an alternate evaluation in which we only score predictions from our literature-augmented models that show a relative confidence increase of at least 10\% over the baseline prediction, presented in Appendix~\ref{sec:highconfinc}. %\an{Ask Lucy for citations for class 2 being medically difficult}
%\subsection{Analyzing Retrieved Literature}

\paragraph{Learning To Retrieve Using Outcomes.}
\beep trains separate models for literature retrieval and outcome prediction. Inspired by \newcite{lee-etal-2019-latent}, we develop a learning-to-retrieve (L2R) formulation that trains both jointly to ensure that the retriever can learn from outcome feedback. However, our L2R model does not improve performance over \beep (results in Table~\ref{tab:l2rresults} in Appendix~\ref{sec:l2r}). We provide discussion for potential reasons in Appendix~\ref{sec:l2r}. This is an interesting direction for future work.

% Given a note $Q$, we first obtain a set of top 100 relevant abstracts ($Docs(Q) = \{D_1,...,D_{100}\}$) from the \beep retrieve-rerank pipeline. The retriever component is then defined as follows:
% \begin{align}
%     E_Q &= BERT_Q(Q)\\
%     E_{D_i} &= BERT_D(D_i)\\
%     S_{retr}(Q, D_i) &= cosine(E_Q, E_{D_i})
% \end{align}\vspace{-0.15em}
% $BERT_Q(X)$ and $BERT_D(X)$ are the query and document encoder functions. Based on retriever scores $S_{retr}$, we select the top $k$ abstracts and perform outcome prediction using the same structure as the \beep outcome prediction module. We also add the following early update loss term to the outcome loss for the retriever component:
% \begin{align}
%     P_{early}(D_i | Q) &= \frac{exp(S_{retr}(Q, D_i))}{\sum_{D_j \in Docs(Q)}exp(S_{retr}(Q, D_j))}\\
%     L_{early} &= - \log{\sum_{D_j \in Docs(Q)} y_j P_{early}(D_j | Q)} 
% \end{align}
% where $y_j$ is set to 1 if using document $D_j$ alongside $Q$ results in a confidence increase in the correct outcome (as per \beep) and 0 otherwise. Our L2R model does not improve performance over \beep (results in Table~\ref{tab:l2rresults} in Appendix~\ref{sec:l2r}). We speculate that this may partly be due to the fact that the heuristic we use to assign $y_j$ values in early update loss is not as accurate as the one used by \newcite{lee-etal-2019-latent} (directly checking for presence of the answer in a document, for the reading comprehension task).
\begin{table*}[t]
\small
    \centering
    \begin{tabular}{L{5.9cm}p{5.1cm}L{2.1cm}L{1.25cm}}
    \toprule \textbf{Patient EHR} & \textbf{Retrieved Abstract} & \textbf{Evidence Type} & \textbf{Outcome}\\ \midrule
    \multirow{1}{5.8cm}{\parbox{5.8cm}{\textbf{CHIEF COMPLAINT:} liver tranplant\\\textbf{PRESENT ILLNESS:} ...s/p liver tranplant...Dx: ESLD secondary to alcoholic \colordg{cirrhosis}.\\\textbf{MEDICAL HISTORY}: EtOH Cirrhosis}} & Retrospective review of data of 73 consecutive patients with \colordg{cirrhosis requiring MV}...majority of patients, 51/64 \colorblue{(79.7\%), dying in the first 48 hours of intubation}...& Patient condition and outcome directly related & No PMV \\ [2mm]
    % & & \\
    % & & \\
    \midrule
    \multirow{1}{5.8cm}{\parbox{5.8cm}{\textbf{CHIEF COMPLAINT: }Aortic dissection\\\textbf{PRESENT ILLNESS: }...72-year-old woman...chest pain...had \colordg{type A aortic dissection}...an intramural hematoma...proceed with \colordg{surgery}...\\\textbf{MEDICAL HISTORY:} HTN Renal failure}}& Acute \colordg{type A aortic dissection} presents a formidable challenge...the most important variables associated with \colorblue{in-hospital mortality} in patients undergoing \colordg{surgery} for this condition...suggests that \colorred{CPB time, diabetes mellitus and postoperative bleeding} are the main determinants of in-hospital death.& Known outcome indicators \textbf{not} present in patient & No MOR \\ [2mm] 
    \midrule
    \multirow{1}{5.8cm}{\parbox{5.3cm}{\textbf{CHIEF COMPLAINT:} Dyspnea, fever\\\textbf{PRESENT ILLNESS:} 58F w/ HCV \colordg{cirrhosis}...requiring BiPAP, ultimately urgent \colordg{intubation}... extubated ... short of breath...\\\textbf{MEDICAL HISTORY:} HCV cirrhosis}} & ...study identifies specific predictors of increased mortality and resource utilization in \colordg{cirrhotic} patients...\colorblue{Increased LOS} in the MICU was \colordg{associated with mechanical ventilation}...
& Ongoing treatment and outcome related & LOS >2 weeks\\

% & & \\
    & & \\
    \bottomrule
    \end{tabular}
    \caption{Qualitative examples of retrieved literature that is 
    % categorized as helpful for cases where adding literature 
    helpful for increasing prediction confidence of the correct outcome. Case 1 shows an example of retrieved literature that strongly matches patient condition and provides direct evidence linking it to the outcome of interest. Case 2 shows an example with indirect evidence, in which retrieved literature lists outcome indicators not present in the patient. Case 3 shows an example of retrieved literature describing a link between patient's ongoing treatment and outcome of interest. \colordg{green}: patient characteristics; \colorblue{blue}: outcome of interest; \colorred{red}: known indicators of the outcome measure not present in the patient.}
    \vspace{-0.85em}
    \label{tab:qualpos}
\end{table*}
\section{Analysis and Discussion}
\label{sec:analysis}
Given \beep's improved performance, we further assess the utility of retrieved literature and cases where adding literature is particularly helpful.

\paragraph{Diversity of retrieved literature.} As a preliminary analysis, we evaluate the diversity of the abstracts retrieved for admission notes in our datasets, as a proxy for the degree to which literature is personalized to specific patient cases. For the 100 most frequently retrieved abstracts for each clinical outcome, Figures~\ref{fig:pmvlitcount},~\ref{fig:morlitcount}, and~\ref{fig:loslitcount} in Appendix \ref{sec:highconfinc} show proportions of patient notes for which these abstracts are judged as relevant by our retrieve-rerank pipeline. From these histograms, we see a stark difference for LOS which is much less diverse than both PMV and MOR, indicating that the literature retrieved for length of stay prediction may be less personalized to patient cases than the literature retrieved for other outcomes. We leave to future work exploration of diversifying retrieved papers across patients and examining the effect on outcome prediction performance.\footnote{We perform an ablation in which we use only the retrieved literature for prediction, showing quantitative evidence for the utility of retrieved literature (see Appendix~\ref{sec:litonly}).}
%This indicates that the literature retrieved for length of stay prediction may be less personalized to patient cases than the literature retrieved for other outcomes, and may partly explain why our models do not show similar improvements on LOS.\\ [-4mm]

\paragraph{Qualitative examination of retrieved literature.} We qualitatively examine literature retrieved for cases in which our model shows large confidence increases over the baseline to determine its utility in making the right prediction. We study increases in both directions, i.e. cases in which adding literature resulted in a confidence increase in either the \emph{correct} outcome label (good) or \emph{incorrect} outcome label (bad). For each clinical outcome, a bio-NLP expert looked at the top 5 cases from each category based on the magnitude of confidence increase (total 10 cases per outcome). For each case, the expert looks at the top 5 abstracts retrieved for the case (total 50 abstracts per outcome) and assigns each abstract to one of 8 categories we define for categorizing degree of relevance and type of evidence provided, including retrievals considered \emph{helpful} and \emph{unhelpful}. For example, see Table~\ref{tab:qualpos} (evidence type column; more in Appendix).

% \begin{enumerate}[leftmargin=*]
% \setlength\itemsep{-0.5em}
%     \item Patient condition and outcome directly related
%     \item Patient history and outcome related
%     \item Known outcome indicators not present in patient
%     \item Ongoing treatment and outcome related
%     \item No cohort match
%     \item No/weak condition match
%     \item Condition-outcome pair not studied
%     \item No evidence for outcome/Weak evidence for direct relationship between patient condition and outcome
% \end{enumerate}

% \begin{figure}
%     \subfloat[Literature categorization for correct outcome cases]{%
%   \includegraphics[clip,scale=0.33]{figures/mor_positive.png}%
% }

% \subfloat[Literature categorization for incorrect outcome cases]{%
%   \includegraphics[clip,scale=0.33]{figures/mor_negative.png}%
% }

% \caption{Literature categorization for mortality. We can see that literature for correct outcome cases is more frequently categorized as helpful, whereas literature for incorrect outcome cases is more frequently unhelpful.}
% \label{fig:morlitcat}
% \end{figure}

\begin{figure}[t!]
    \centering
    \includegraphics[scale=0.24]{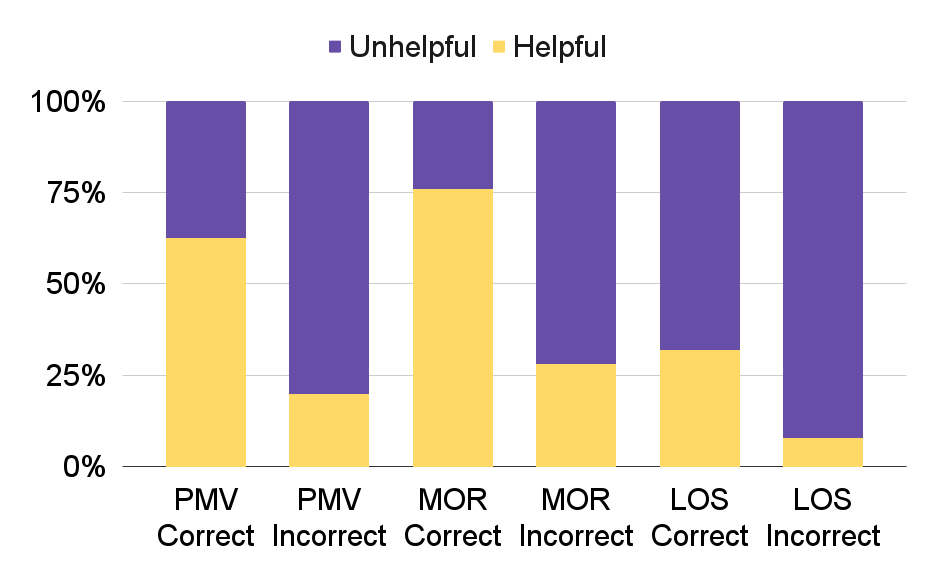}
    \caption{Literature categorization for both correct and incorrect outcome cases. For PMV and MOR, retrieved literature for correct cases is more often categorized as helpful, and unhelpful literature dominates for incorrect cases. For LOS, literature for both categories is more often categorized as unhelpful.}
    \label{fig:litutil}
    \vspace{-1em}
\end{figure}

As seen in Table~\ref{tab:qualpos}, for \emph{helpful} categories, retrieved literature matches patient characteristics (especially current condition) and includes evidential links between outcome of interest and patient conditions/treatment. In the first case, the retrieved abstract provides evidence that patients with cirrhosis have high mortality in the first 48 hours of intubation, entails the patient might not undergo prolonged ventilation. In the second case, the abstract lists comorbidities associated with in-hospital mortality (outcome of interest), but none are present in the patient under consideration, which can be taken as weak indication that the patient may survive. Similarly, for the third case, the retrieved abstract mentions that cirrhotic patients may have longer hospital stays if they are on mechanical ventilation. This matches our patient's treatment history since she has cirrhosis and was briefly intubated and extubated, before experiencing shortness of breath again. Given this, the patient might have a longer length of stay. Conversely, \emph{unhelpful} retrieved literature often does not match patient characteristics or may not contain evidence relevant to the outcome. See more example explanations in Appendix~\ref{sec:qualneg}. 

Figure~\ref{fig:litutil} presents the distribution of helpful and unhelpful categories for both kinds of cases for all outcomes. We can see that for correct outcome cases from both PMV and mortality, retrieved literature is more frequently assigned to one of the helpful categories, while for incorrect outcome cases, retrieved literature is more frequently assigned to one of the unhelpful categories. For LOS, unhelpful categories dominate both types of cases, especially prevalent in incorrect outcomes. 

%We see similar trends for prolonged mechanical ventilation, but mixed results for length of stay predictions (graphs in Appendix~\ref{sec:litcat}).

% \textbf{Utility of literature for UMLSBERT vs BLUEBERT:} \\

% \subsection{Learning To Retrieve Using Outcome Feedback?}
\section{Conclusion}

In this paper, we introduced \beep, a system that automatically retrieves \emph{patient-specific literature} based on intensive care (ICU) EHR notes and uses the literature to enhance clinical outcome prediction. On three challenging tasks, we obtain substantial improvements over strong recent baselines, seeing dramatic gains in top-10\% precision for mortality prediction with a boost of over 25\%.

Our hope is that this work will open new research directions into bridging the gap between AI-based clinical models and the Evidence Based Medicine (EBM) paradigm in which medical decisions are based on explicit evidence from the literature. An interesting direction is to incorporate evidence identification and inference \cite{wallace2019does,deyoung2020evidence} directly into our retrieval and predictive models. Another important question to explore relates to the implications our approach has on increasing the interpretability of clinical AI models.

\section*{Acknowledgements}
This work was supported in part by NSF Convergence Accelerator Award \#2132318. The authors would like to thank the members of the Semantic Scholar team, and the anonymous reviewers for their helpful feedback on this work.

\section*{Ethical Concerns}
Incorporating outcome prediction models into a medical decision-making pipeline effectively will require these technologies to adhere to standards set by the core principles of medical ethics: beneficence, non-maleficence, autonomy, and justice \cite{beauchamp2001principles}. These requirements may raise the following concerns when deploying outcome prediction models in clinical settings:
\begin{itemize}[topsep=0pt,leftmargin=*]
\setlength\itemsep{-0.5em}
    \item \textbf{Out-of-Cohort Generalization:} The extent to which outcome prediction models generalize to patient cohorts that may not have been present in their training data is unclear. If model accuracy is significantly lower on ``out-of-cohort'' patients, using inaccurate/uncertain predictions during decision making may violate the requirement that any application of technology must be \textit{beneficent} and \textit{non-maleficent} to individual patients. Our proposed technique can partly mitigate the generalization issue by identifying additional supporting evidence from literature, which may be better tailored to individual patient characteristics, instead of using only cohort-level evidence. However, biomedical literature can also have blind spots, with certain cohorts and disease combinations being under-studied, and even literature-augmented prediction may not be sufficiently accurate.
    \item \textbf{Algorithmic Biases:} Since outcome prediction models are trained on historical health data, existing inequities in healthcare access may translate into models continuing to perpetuate unintentional discrimination against patients from under-served demographics. For example, models might predict poorer outcomes (e.g., high mortality, poor response to treatment, etc.) for specific demographics that have historically had worse outcomes due to poor access to care. Such issues are a clear violation of the \textit{justice} requirement, and must be tackled before deployment.
    % can lead to unintentional discrimination against patients from specific demographics, violating the principle of \textit{justice}. Systemic inequities Such discrimination may stem from various causes.  
    \item \textbf{Informed Consent:} Lastly, if outcome prediction models are used in clinical settings, patients and their caregivers must be made aware of their use, since the principle of \textit{autonomy} emphasizes that patients must be provided all relevant medical information to support autonomous decision making. The black-box nature of these models raises another issue: how can we help patients/caregivers understand and interpret outcome predictions to further support their autonomy in decision making? We hope that literature-augmented prediction techniques can partly ease this by using evidence snippets from literature that contributed to the model's prediction as explanations. 
    % \item \textbf{Lack of Transparency:}
\end{itemize}

\bibliography{custom}
\bibliographystyle{acl_natbib}

\appendix
\section{PubMed Queries Per Outcome}
\label{sec:pubquery}
Following are the MeSH terms that we use to retrieve literature from PubMed to construct the outcome-specific index for each clinical outcome under consideration:
\begin{itemize}
    \item \textbf{Prolonged Mechanical Ventilation (PMV):} ``Respiration, Artificial''. We also query using the terms ``Ventilation, Mechanical'' and ``Ventilator Weaning'' but do not find any new results.
    \item \textbf{In-Hospital Mortality (MOR):} ``Hospital Mortality'', ``Mortality+Humans+Risk Factors''. Note that the ``+'' operator is interpreted as AND by PubMed search.
    \item \textbf{Length of Stay (LOS):} ``Length of Stay''. All other MeSH terms from the tagger are aliases of this term.
\end{itemize}

\begin{table*}
    \centering
    \footnotesize
    \begin{tabular}{lcccccc}
     \toprule & \multicolumn{2}{c}{PMV} & \multicolumn{2}{c}{MOR} & \multicolumn{2}{c}{LOS} \\ \cmidrule{2-7}
     \textbf{LM} & \textbf{AUROC} & \textbf{Micro F1 } & \textbf{AUROC} & \textbf{Micro F1 } & \textbf{AUROC} & \textbf{Micro F1 } \\ \midrule
    \textbf{ClinicalBERT (Full)} & 54.66 & 53.93 & 81.78 & 86.34 & 70.94 & 40.00 \\ 
    \textbf{ClinicalBERT (Disc.)} & 54.91 & 54.21 & 81.78 & 86.34 & 71.44 & 40.36\\
    \textbf{CORe} & 54.98 & 54.35 & 81.58 & 84.85 & 69.15 & 37.94 \\ 
    \textbf{BLUEBERT} & 56.60 & 55.34 & 82.40 & 84.75 & \textbf{71.87} & \textbf{41.93} \\ 
    \textbf{UMLSBERT} & \textbf{57.42} & \textbf{55.48} & \textbf{83.31} & \textbf{87.29} & 71.60 & 41.84 \\ \bottomrule
    \end{tabular}
    \caption{Performance of various language models trained on clinical and biomedical text on all clinical outcomes. For ClinicalBERT, Disc.~and Full refer respectively to variants trained on discharge summaries only and both discharge summaries and nursing notes.}
    \label{tab:lm}
\end{table*}

\section{Implementation Details}
\label{sec:imp}

\noindent \textbf{Entity Extraction.} First, we extract entities from clinical notes using a model trained on the i2b2 2010 concept extraction dataset \cite{uzuner20112010}. This dataset consists of clinical notes annotated with three types of entities: problems, tests, and treatments. These entity types cover the pertinent medical information that can be used to retrieve abstracts relevant to a clinical note. Moreover, the i2b2 guidelines require annotators to include all qualifiers within an entity span, so training a model on these annotations should bias it towards including pertinent entity qualifiers. Our entity extraction model uses a BERT-based language model to compute token representations, followed by a linear layer to predict entity labels.\\ [-4mm]

We use ClinicalBERT \cite{alsentzer-etal-2019-publicly} as the the language model to train our i2b2 entity extractor. Table~\ref{tab:i2b22010} shows the performance of our model on the i2b2 2010 test set. These numbers are close to the exact F1 scores reported by \newcite{alsentzer-etal-2019-publicly} on i2b2 2010 (87.8).
\begin{table}[h]
    \centering
    \begin{tabular}{lc}
        \toprule \textbf{Category} & \textbf{Exact F1} \\ \midrule
        Overall & 86.66 \\
        Test & 87.48 \\
        Problem & 86.53 \\
        Treatment & 86.03 \\ \bottomrule
    \end{tabular}
    \caption{Entity extraction model performance on i2b2 2010 test set}
    \label{tab:i2b22010}
\end{table}

\noindent \textbf{Entity Filtering.} After extracting entities, we filter out all negated entities. Negated entities are detected using the ConText algorithm for negation detection from clinical text \cite{harkema2009context}. We use the implementation of ConText negated entity detection algorithm provided by medspaCy \cite{eyre2021medspacy}.\\

\noindent \textbf{MeSH Linking.} Finally, the set of filtered entities is linked to MeSH terms using scispaCy. Entities not linked to MeSH terms are discarded. MeSH terms linked in clinical notes and abstracts are used to compute TF-IDF vectors for the sparse retrieval model.

\noindent
\textbf{Bi-Encoder} 
Given a query clinical note $Q$ and an abstract document $D_i$, a BERT-based encoder is used to compute dense embedding representations $E_Q$ and $E_{D_i}$. A scoring function $S$ is defined as the Euclidean distance between query and document embeddings:
\begin{equation}
%\small
  S(Q, D_i) = \lVert E_Q - E_{D_i} \rVert_2 
\end{equation}
Documents closest to the query vector in the embedding space are returned as top-ranked results. The bi-encoder is trained using a triplet loss function defined as follows:
\begin{multline}
  L(Q, D_i^+, D_i^-) = \\
  \max(S(Q, D_i^+) - S(Q, D_i^-) + m, 0)
\end{multline}
Here $D_i^+$ is an abstract more relevant to the clinical note $Q$ than $D_i^-$ and $m$ is a margin value. We use PubmedBERT \cite{gu2021domain} as the encoder and train our bi-encoder using the dataset from the TREC 2016 clinical decision support task \cite{DBLP:conf/trec/RobertsDVH16}.\footnote{We do not use data from TREC 2014 and 2015 since they use idealized case reports instead of actual EHR notes. Combining all three datasets degraded performance, likely due to differences in language between case reports and EHRs.} This dataset consists of 30 de-identified EHR notes, along with $\sim$1000 PubMed abstracts per note marked for relevance. We select relevant abstracts per note as positive candidates ($D_i^+$), and irrelevant abstracts for the same note as negative candidates ($D_i^-$).

\noindent
\textbf{Outcome prediction module training.} We use a weighted cross-entropy loss function to handle class imbalance. Given a dataset with $N$ total examples, $c$ classes and $n_i$ examples in class $i$, class weights are computed as follows:
\begin{equation}
    w_i = \frac{N}{c \cdot n_i}
\end{equation}
We use Adam optimizer, treating initial learning rate as a hyperparameter. All models are implemented in PyTorch, and we use Huggingface implementations for all pretrained language models.

\begin{table*}[]
    \centering
    \small
    \setlength{\tabcolsep}{4.5pt}
    \begin{tabular}{lccccccccc}
    \toprule  & \multicolumn{3}{c}{PMV} & \multicolumn{3}{c}{MOR} & \multicolumn{3}{c}{LOS} \\ \cmidrule{2-10}
    \textbf{Model} & \textbf{AUROC} & \textbf{Micro F1} & \textbf{Macro F1} & \textbf{AUROC} & \textbf{Micro F1} & \textbf{Macro F1} & \textbf{AUROC} & \textbf{Micro F1} & \textbf{Macro F1} \\ \midrule
    \textbf{UMLSBERT} & 56.44 & 56.07 & {54.97} & 83.34 & 87.93 & 66.93 & 72.19 & 43.12 & 42.20 \\
    \textbf{\qquad+Avg} & 54.17 & 53.53 & 41.51 & 84.54 & 90.47 & 60.53 & 71.90 & 44.88 & 41.26 \\
    \textbf{\qquad+SVote} & 54.29 & 52.82 & 39.93 & 84.50 & 90.51 & 61.10 & 72.17 & 45.56 & 41.68 \\
    \textbf{\qquad+WVote} & 57.60 & \textbf{56.50} & \textbf{55.93} & 83.92 & 90.54 & 61.20 & 72.72 & 46.46 & 42.17 \\
    \textbf{\qquad+WAvg} & \textbf{58.65} & 55.79 & 53.68 & 84.68 & 90.59 & 62.78 & 72.16 & 45.04 & 40.87 \\
\bottomrule
    \end{tabular}
    \caption{Performance of learning to retrieve (L2R) model on all clinical outcomes using the UMLSBERT language model}
    \label{tab:l2rresults}
\end{table*}

\begin{table*}[]
    \centering
    \small
    \setlength{\tabcolsep}{4.5pt}
    \begin{tabular}{lccccccccc}
    \toprule  & \multicolumn{3}{c}{PMV} & \multicolumn{3}{c}{MOR} & \multicolumn{3}{c}{LOS} \\ \cmidrule{2-10}
    \textbf{Model} & \textbf{AUROC} & \textbf{Micro F1} & \textbf{Macro F1} & \textbf{AUROC} & \textbf{Micro F1} & \textbf{Macro F1} & \textbf{AUROC} & \textbf{Micro F1} & \textbf{Macro F1} \\ \midrule
    \textbf{CORe} & 55.91 & 53.96 & 53.71 & 79.96 & 78.92 & 62.46 & 71.52 & 42.59 & \textbf{42.33} \\
    \textbf{\qquad+Avg} & \textbf{58.76} & 55.51 & 55.43 & 82.41 & 84.67 & \textbf{66.06} & \textbf{71.99} & 40.54 & 40.39 \\
    \textbf{\qquad+SVote} & 58.40 & \textbf{58.62} & 55.23 & 81.90 & \textbf{89.90} & 55.76 & 71.35 & \textbf{45.07} & 40.16 \\
    \textbf{\qquad+WVote} & 58.03 & 56.92 & 53.14 & \textbf{82.81} & \textbf{89.87} & 53.16 & 70.96 & 44.74 & 39.73 \\
    \textbf{\qquad+WAvg} & 57.53 & 55.51 & \textbf{55.49} & 81.98 & 81.86 & 64.63 & 71.17 & 39.48 & 39.67 \\
\bottomrule
    \end{tabular}
    \caption{Performance of baseline and literature-augmented outcome prediction models on all clinical outcomes using the CORe language model}
    \label{tab:resultscore}
\end{table*}

\begin{table*}[]
    \centering
    \small
    \setlength{\tabcolsep}{4.5pt}
    \begin{tabular}{lccccccccc}
    \toprule  & \multicolumn{3}{c}{PMV} & \multicolumn{3}{c}{MOR} & \multicolumn{3}{c}{LOS} \\ \cmidrule{2-10}
    \textbf{Model} & \textbf{AUROC} & \textbf{Micro F1} & \textbf{Macro F1} & \textbf{AUROC} & \textbf{Micro F1} & \textbf{Macro F1} & \textbf{AUROC} & \textbf{Micro F1} & \textbf{Macro F1} \\ \midrule
    \textbf{BLUEBERT} & -- & -- & -- & -- & -- & -- & -- & -- & --\\
    \textbf{\qquad+Avg} & 55.72 & 54.38 & 46.95 & 68.72 & 89.49 & 47.23 & 63.40 & 39.40 & 29.15 \\
    \textbf{\qquad+SVote} & 57.11 & 56.50 & 52.21 & 71.04 & 89.49 & 48.73 & 63.46 & 39.41 & 28.90 \\
    \textbf{\qquad+WVote} & 55.83 & 53.25 & 43.43 & 71.00 & 89.50 & 48.73 & 63.40 & 39.56 & 27.52 \\
    \textbf{\qquad+WAvg} & 56.99 & 55.65 & 47.97 & 71.39 & 89.48 & 49.26 & 63.46 & 39.34 & 27.99 \\ \midrule
    \textbf{UMLSBERT} & -- & -- & -- & -- & -- & -- & -- & -- & --\\
    \textbf{\qquad+Avg} & 59.15 & 55.37 & 50.79 & 71.22 & 89.49 & 48.54 & 63.84 & 39.49 & 30.30 \\
    \textbf{\qquad+SVote} & 56.53 & 55.09 & 51.76 & 69.31 & 89.50 & 47.71 & 63.14 & 38.95 & 27.12 \\
    \textbf{\qquad+WVote} & 57.06 & 54.38 & 53.77 & 70.54 & 89.46 & 49.34 & 63.46 & 39.40 & 27.55 \\
    \textbf{\qquad+WAvg} & 56.99 & 54.94 & 54.29 & 70.04 & 89.46 & 49.16 & 63.51 & 39.51 & 28.32 \\
\bottomrule
    \end{tabular}
    \caption{Performance of models that only use retrieved literature for outcome prediction on all clinical outcomes}
    \label{tab:litonly}
\end{table*}

% \begin{table*}[]
%     \centering
%     \small
%     \setlength{\tabcolsep}{4.5pt}
%     \begin{tabular}{lccccccccc}
%     \toprule  & \multicolumn{3}{c}{PMV} & \multicolumn{3}{c}{MOR} & \multicolumn{3}{c}{LOS} \\ \cmidrule{2-10}
%     \textbf{Model} & \textbf{AUROC} & \textbf{Micro F1} & \textbf{Macro F1} & \textbf{AUROC} & \textbf{Micro F1} & \textbf{Macro F1} & \textbf{AUROC} & \textbf{Micro F1} & \textbf{Macro F1} \\ \midrule
%     \textbf{BLUEBERT} & -- & -- & -- & -- & -- & -- & -- & -- & --\\
%     \textbf{\qquad+Avg} &  &  &  &  &  &  & 71.45 & 44.46 & 39.78 \\
%     \textbf{\qquad+SVote} &  &  &  &  &  &  & 72.35 & 45.94 & 42.14 \\
%     \textbf{\qquad+WVote} &  &  &  &  &  &  &  &  &  \\
%     \textbf{\qquad+WAvg} &  &  &  &  &  &  &  &  &  \\ \midrule
%     \textbf{UMLSBERT} & -- & -- & -- & -- & -- & -- & -- & -- & --\\
%     \textbf{\qquad+Avg} &  &  &  &  &  &  &  &  &  \\
%     \textbf{\qquad+SVote} &  &  &  &  &  &  &  &  &  \\
%     \textbf{\qquad+WVote} &  &  &  &  &  &  &  &  &  \\
%     \textbf{\qquad+WAvg} &  &  &  &  &  &  &  &  &  \\
% \bottomrule
%     \end{tabular}
%     \caption{Performance of models on all clinical outcomes when retrieved literature is randomly shuffled at test time}
%     \label{tab:litonly}
% \end{table*}

\begin{table*}[]
\small
    \centering
    \begin{tabular}{lcccccccc}
     \toprule & \multicolumn{2}{c}{\textbf{<3 days}} & \multicolumn{2}{c}{\textbf{>=3 and <=7 days}} & \multicolumn{2}{c}{\textbf{>7 and <=14 days}} & \multicolumn{2}{c}{\textbf{>14 days}}\\ \cmidrule{2-9}
     \textbf{Model} & \textbf{Prec@10} & \textbf{Rec@10} & \textbf{Prec@10} & \textbf{Rec@10} & \textbf{Prec@10} & \textbf{Rec@10} & \textbf{Prec@10} & \textbf{Rec@10}\\ \midrule
    \textbf{BLUEBERT} & 47.6 & 37.11 & 61.09 & 16.14 & 44.98 & 14.15 & 50.74 & 26.93 \\
    \textbf{\qquad+Avg} & 54.23 & 24.0 & 60.64 & 16.02 & 45.45 & 14.49 & 49.48 & 25.66 \\
    \textbf{\qquad+SVote} & 54.48 & 27.12 & 62.12 & 16.41 & 46.38 & 14.97 & 51.33 & 26.87 \\
    \textbf{\qquad+WVote} & 55.73 & 21.68 & 61.66 & 16.29 & 46.68 & 12.78 & 47.99 & 25.18 \\
    \textbf{\qquad+WAvg} & 52.48 & 28.28 & 60.75 & 16.05 & 47.33 & 15.12 & 51.03 & 26.99 \\ \midrule
    \textbf{UMLSBERT} & 47.33 & 37.11 & 59.95 & 15.84 & 44.83 & 13.04 & 48.92 & 25.97\\
    \textbf{\qquad+Avg} & 53.08 & 26.14 & 60.41 & 15.96 & 48.3 & 15.27 & 49.6 & 26.03 \\
    \textbf{\qquad+SVote} & 52.37 & 28.55 & 59.5 & 15.72 & 44.38 & 14.38 & 49.36 & 25.72 \\
    \textbf{\qquad+WVote} & 57.22 & 27.21 & 64.28 & 16.98 & 45.43 & 14.78 & 50.4 & 26.33 \\
    \textbf{\qquad+WAvg} & 52.86 & 20.61 & 59.84 & 15.81 & 44.9 & 14.38 & 48.44 & 25.24 \\
\bottomrule     
    \end{tabular}
    \caption{Precision and recall scores for top 10\% high-confidence predictions per class (precision/recall@TOP-K) for LOS.}
    \label{tab:lostop}
\end{table*}

\section{Hyperparameter Tuning}
\label{sec:tuning}
We do a grid search over the following hyperparameter values for each aggregation:\\
\noindent
\textbf{Learning Rate (LR): }[5e-4, 1e-5, 5e-5, 1e-6,5e-6]\\
\noindent
\textbf{Number of top abstracts (k):} [1, 5, 10]\\
\noindent
\textbf{Gradient accumulation steps (GA):} [10, 20]
This hyperparameter grid stays consistent across all outcome prediction experiments. For all experiments, we currently report the outcome of a single run.

\section{Computing Infrastructure}
Our experiments were carried out on 2 AWS p3.16xlarge instances, which are 8-GPU machines with 16 GB RAM per GPU. All our experiments can be run on a single 16 GB GPU.

\section{Results from Learning To Retrieve Model}
\label{sec:l2r}

Given a note $Q$, we first obtain a set of top 100 relevant abstracts ($Docs(Q) = \{D_1,...,D_{100}\}$) from the \beep retrieve-rerank pipeline. The retriever component is then defined as follows:
\begin{align}
    E_Q &= BERT_Q(Q)\\
    E_{D_i} &= BERT_D(D_i)\\
    S_{retr}(Q, D_i) &= cosine(E_Q, E_{D_i})
\end{align}\vspace{-0.15em}
$BERT_Q(X)$ and $BERT_D(X)$ are the query and document encoder functions. Based on retriever scores $S_{retr}$, we select the top $k$ abstracts and perform outcome prediction using the same structure as the \beep outcome prediction module. We also add the following early update loss term to the outcome loss for the retriever component:
\begin{align}
    P_{early}(D_i | Q) &= \frac{exp(S_{retr}(Q, D_i))}{\sum_{D_j \in Docs(Q)}exp(S_{retr}(Q, D_j))}\\
    L_{early} &= - \log{\sum_{D_j \in Docs(Q)} y_j P_{early}(D_j | Q)} 
\end{align}
where $y_j$ is set to 1 if using document $D_j$ alongside $Q$ results in a confidence increase in the correct outcome (as per \beep) and 0 otherwise. Our L2R model does not improve performance over \beep (results in Table~\ref{tab:l2rresults}). We speculate that this may partly be due to the fact that the heuristic we use to assign $y_j$ values in early update loss is not as accurate as the one used by \newcite{lee-etal-2019-latent} (directly checking for presence of the answer in a document, for the reading comprehension task).

Table~\ref{tab:l2rresults} presents results for the learning-to-retrieve model on all clinical outcomes using UMLSBERT as the encoder. From the table, we can see that while L2R improves performance over a notes-only baseline, its performance is comparable to \beep. As mentioned earlier, we speculate that this may partly be attributed to the fact that the heuristic we use to assign $y_j$ values in early update loss is not as accurate as the one used by \newcite{lee-etal-2019-latent} (directly checking for presence of answer in document, for the reading comprehension task). We believe that experimenting with other sources of supervision to generate $y_j$ values and weighting mechanisms to better combine outcome and early update losses might lead to larger improvements, but we leave those to future work.

\section{Literature-Augmented Outcome Prediction with CORe}
\label{sec:core}
Table~\ref{tab:resultscore} shows the overall performance of our literature-augmented outcome prediction system on all three clinical outcomes when the CORe language model is used as an encoder. From this table, we can see that adding literature improves performance in this setting as well (with the exception of macro F1 on length of stay). However the overall scores are lower than the settings in which UMLSBERT and BLUEBERT are used as encoders (Table~\ref{tab:results}).

\section{Literature-Only Outcome Prediction}
\label{sec:litonly}
To quantitatively test the quality of the retrieved literature, we run an ablation study in which we predict the clinical outcome using only the literature retrieved for a specific patient case, without incorporating any information from the patient clinical note. Table~\ref{tab:litonly} shows the results for this ablation study, using both BLUEBERT and UMLSBERT encoders. From this table, we can see that while removing the clinical note leads to performance drops, especially on mortality and length of stay, the retrieved literature does have some predictive ability. We take this as indication that the retrieved literature contains some clinical indicators associated with the outcome, that are also present in the patient's clinical note. 

\section{Analyzing High Confidence Increases Over Baseline}
\label{sec:highconfinc}

\begin{table}[]
\begin{subtable}{1\columnwidth}
\small
\scalebox{0.9}{
    \centering
    \begin{tabular}{lcc}
     \toprule \textbf{Model} & \textbf{No PMV} & \textbf{PMV} \\ \midrule
    \textbf{BLUEBERT+Avg} & 55.47 & 57.48 \\
    \textbf{BLUEBERT+SVote} & 56.82 & 55.56 \\
    \textbf{BLUEBERT+WVote} & 62.50 & 62.67 \\
    \textbf{BLUEBERT+WAvg} & 56.34 & 61.29 \\ \midrule
    \textbf{UMLSBERT+Avg} & 63.71 & 60.71 \\
    \textbf{UMLSBERT+SVote} & 50.39 & 65.62 \\
    \textbf{UMLSBERT+WVote} & 61.83 & 59.09 \\
    \textbf{UMLSBERT+WAvg} & 57.80 & 63.33 \\
\bottomrule     
    \end{tabular}
}
    \caption{Precision on PMV, when considering cases for which literature-augmented models achieve >10\% increase in prediction confidence over baseline.}
    \label{tab:pmvconf}
\end{subtable}
\begin{subtable}{1\columnwidth}
\small
\scalebox{0.9}{
    \centering
    \begin{tabular}{lcc}
     \toprule \textbf{Model} & \textbf{No MOR} & \textbf{MOR} \\ \midrule
    \textbf{BLUEBERT+Avg} & 87.91 & 69.77 \\
    \textbf{BLUEBERT+SVote} & 87.49 & 75.00 \\
    \textbf{BLUEBERT+WVote} & 86.99 & 76.09 \\
    \textbf{BLUEBERT+WAvg} & 87.29 & 77.68 \\ \midrule
    \textbf{UMLSBERT+Avg} & 85.33 & 83.33 \\
    \textbf{UMLSBERT+SVote} & 90.33 & 31.01 \\
    \textbf{UMLSBERT+WVote} & 86.66 & 52.17 \\
    \textbf{UMLSBERT+WAvg} & 85.29 & 60.00 \\
\bottomrule     
    \end{tabular}
}
    \caption{Precision on MOR, when considering cases for which literature-augmented models achieve >10\% increase in prediction confidence over baseline.}
    \label{tab:morconf}
\end{subtable}
\end{table}

Finally, we also examine an alternate way of using high-confidence predictions made by our models. We run both baseline and literature-augmented systems, and only consider predictions from the literature-augmented system that show a high increase in confidence, such as $>10\%$ increase relative to the baseline predictions for the same cases. Tables~\ref{tab:pmvconf} and~\ref{tab:morconf} show the precision scores of all models on prolonged mechanical ventilation and mortality in this setting. 
% For this evaluation we only consider cases for which literature-augmented models show a confidence increase greater than 10\%. 
We can see that precision scores in this setting are fairly high, especially for the negative class in mortality prediction. Most averaging variants also do well on the positive class in mortality prediction.

\section{Examples of Literature For Incorrect Outcome Cases}
\label{sec:qualneg}

We categorize examples into the following:
\begin{enumerate}[leftmargin=*]
\setlength\itemsep{-0.5em}
    \item Patient condition and outcome directly related
    \item Patient history and outcome related
    \item Known outcome indicators not present in patient
    \item Ongoing treatment and outcome related
    \item No cohort match
    \item No/weak condition match
    \item Condition-outcome pair not studied
    \item No evidence for outcome/Weak evidence for direct relationship between patient condition and outcome
\end{enumerate}

\begin{table*}[h]
\small
    \centering
    \begin{tabular}{p{5cm}p{5cm}p{3.5cm}p{1.5cm}}
    \toprule \textbf{Patient EHR} & \textbf{Retrieved Abstract} & \textbf{Evidence Type} & \textbf{Outcome}\\ \midrule
    \multirow{2}{5cm}{\parbox{5cm}{\textbf{CHIEF COMPLAINT:} ICD firing\\\textbf{PRESENT ILLNESS:} 57 yo M presenting s/p ICD discharges...shocks preceded by prodrome of dizziness,...and was shocked once...Has not had ICD firing prior to these events since implant\\\textbf{MEDICAL HISTORY:} Heart failure...}} & ...assess if selected clinical markers of organ dysfunction were associated with increased 1-year mortality despite ICD therapy...Clinical markers of liver dysfunction, recent mechanical ventilation, and renal impairment were independently associated with increased 1 year mortality...& Weak condition match, condition-outcome pair not studied & PMV\\ \midrule
    \multirow{2}{5cm}{\parbox{5cm}{\textbf{CHIEF COMPLAINT:} acute onset right hemiplegia and aphasia\\\textbf{PRESENT ILLNESS:} 84yo M...acute onset of inability to speak and right hemiplegia...head CT showed dense L MCA and hypodensities in left inferior frontal lobe and left corona radiata.\\\textbf{MEDICAL HISTORY:} HTN Afib, off coumadin...}}& Stroke is indicated by an abrupt manifestation of neurologic deficits secondary to an ischemic or hemorrhagic insult to a region of the brain...ranked as the third leading cause of death in the United States...report shows that despite the use of antithrombotic and/or antiplatelet aggregating drugs, the key to stroke management is primary prevention. & No cohort match, condition-outcome pair not studied & MOR\\ \midrule
    \multirow{2}{5cm}{\parbox{5cm}{\textbf{CHIEF COMPLAINT:} Substernal chest pain\\\textbf{PRESENT ILLNESS:} ...62 yo M... no prior cardiac history... substernal CP... mild SOB, nausea, diaphoresis and numbness in left arm... \\\textbf{MEDICAL HISTORY:} foot surgery 2 weeks ago ?COPD ?gastritis? }} & ..rising health care costs have created pressures to increase efficiency of coronary care units. Possible strategies seek to decrease resource use by identifying low-risk patients for initial triage or early transfer to lower levels of care... & No cohort match, no evidence for outcome & LOS <3 days\\
& & \\
& & \\
     \bottomrule
    \end{tabular}
    \caption{Qualitative examples of retrieved literature that is categorized as unhelpful for cases where adding literature increases confidence in incorrect outcome. Case 1 shows an example of retrieved literature that has a weak match with patient condition, but no evidence linking condition to outcome. Case 2 shows an example in which retrieved literature does not match patient case or contain evidence for outcome. Case 3 shows an example of a review article that again does not match patient case or provide outcome evidence.}
    \label{tab:qualneg}
\end{table*}

\begin{figure*}[t!]
     \centering
     \begin{subfigure}{0.3\textwidth}
         \centering
         \includegraphics[width=\textwidth]{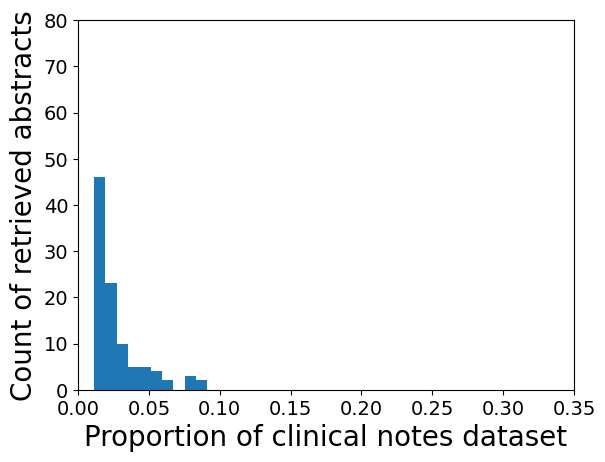}
         \caption{PMV}
         \label{fig:pmvlitcount}
     \end{subfigure}
     \hfill
     \begin{subfigure}{0.3\textwidth}
         \centering
         \includegraphics[width=\textwidth]{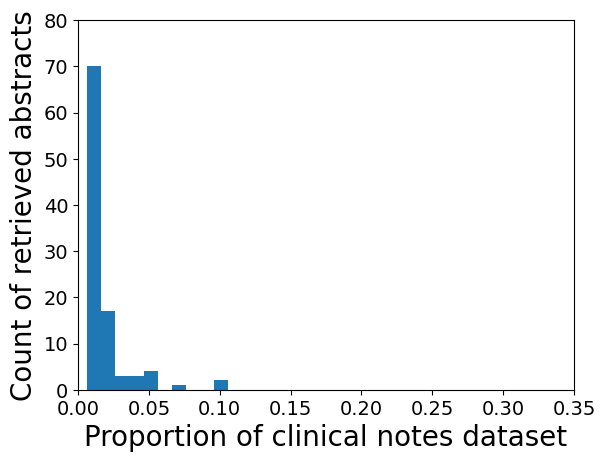}
         \caption{MOR}
         \label{fig:morlitcount}
     \end{subfigure}
     \hfill
     \begin{subfigure}{0.3\textwidth}
         \centering
         \includegraphics[width=\textwidth]{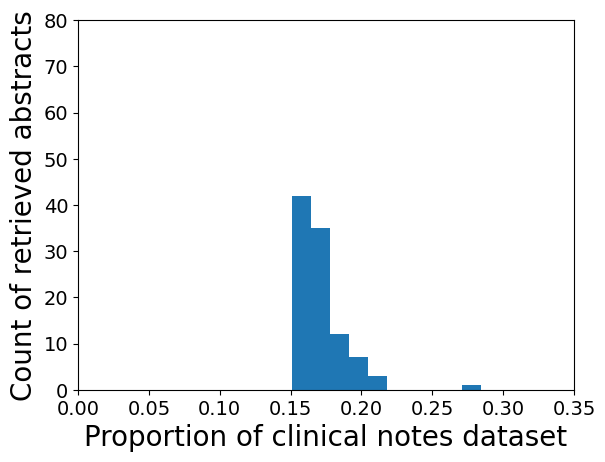}
         \caption{LOS}
         \label{fig:loslitcount}
     \end{subfigure}
     \caption{Proportion of admission notes associated with the 100 most highly retrieved abstracts for each clinical outcome. From these graphs, we can see that frequently-retrieved abstracts for LOS are associated with a larger proportion of cases from the dataset, than frequently retrieved abstracts for PMV and MOR (indicative of lower literature diversity in LOS).}
    %\vspace{-3mm}
\end{figure*}

From table~\ref{tab:qualneg}, we can see that retrieved literature from \emph{unhelpful} categories often does not match patient characteristics. The first case discusses a patient who has had an ICD firing incident, but the retrieved literature discusses ICD implantation therapy. While related, there is no discussion of the impact of ICD firing on various clinical outcomes. 

For the second case, we see that the retrieved article discusses strokes in general, without matching any of the patient's indications or demographic characteristics. Moreover, the outcome of interest (mortality) is mentioned briefly, but links between the outcome and patient conditions are not studied. Finally, the third case provides an example of a common phenomenon we observe. There are a fair number of review articles retrieved that do not have strong evidential statements in the abstract. For the third case, the retrieved abstract discusses the need for early triage/transfer (which could lead to low length of stay), but then do not provide any conclusive evidence.

\end{document}